\documentclass[conference]{IEEEtran}
\usepackage{times}

\usepackage[numbers]{natbib}
\usepackage{multicol}
\usepackage{hyperref}

\usepackage[pdftex]{graphicx}
\usepackage{subfig}
\usepackage{amsmath}
\usepackage{amssymb}
\usepackage{bm}
\include{natbib.sty}
\usepackage{cleveref}
\usepackage{acro}
\usepackage{relsize}
\usepackage{siunitx}
\usepackage{threeparttable}
\acsetup{short-format=\textsmaller}
\usepackage{balance}

\IEEEoverridecommandlockouts

\DeclareAcronym{NDT}{short = NDT, long = non-destructive testing}
\DeclareAcronym{DOF}{short = DoF, long = degrees of freedom}
\DeclareAcronym{MAV}{short = MAV, long = micro aerial vehicle}
\DeclareAcronym{OMAV}{short = OMAV, long = omnidirectional micro aerial vehicle}
\DeclareAcronym{AM}{short = AM, long = aerial manipulator}
\DeclareAcronym{UAV}{short = UAV, long = unmanned aerial vehicle}
\DeclareAcronym{GPS}{short = GPS, long = global positioning system}
\DeclareAcronym{IMU}{short = IMU, long = inertial measurement unit}
\DeclareAcronym{TOF}{short = ToF, long = time-of-flight}
\DeclareAcronym{VI}{short = VI, long = visual-inertial}
\DeclareAcronym{CSE}{short = CSE, long = copper sulfate electrode}
\DeclareAcronym{PID}{short = PID, long = proportional-integral-derivative}
\DeclareAcronym{PD}{short = PD, long = proportional-derivative}
\DeclareAcronym{RMSE}{short = RMSE, long = root mean squared error}

\begin{document}

% paper title
\title{An Omnidirectional Aerial Manipulation Platform for Contact-Based Inspection}

\author{\IEEEauthorblockN{
Karen Bodie\IEEEauthorrefmark{1},
Maximilian Brunner\IEEEauthorrefmark{1},
Michael Pantic\IEEEauthorrefmark{1},
Stefan Walser\IEEEauthorrefmark{1},
Patrick Pf\"{a}ndler\IEEEauthorrefmark{2}, \\
Ueli Angst\IEEEauthorrefmark{2},
Roland Siegwart\IEEEauthorrefmark{1} and
Juan Nieto\IEEEauthorrefmark{1}}
\thanks{\IEEEauthorrefmark{1}\href{http://www.asl.ethz.ch/}{Autonomous Systems Lab}, ETH Z\"{u}rich, Switzerland}
\thanks{\IEEEauthorrefmark{2}\href{http://www.ifb.ethz.ch/}{Institute for Building Materials}, ETH Z\"{u}rich, Switzerland}
\thanks{\: e-mail: \href{mailto:kbodie@ethz.ch}{\texttt{kbodie@ethz.ch}} }
\\[-3.0ex]
}

\maketitle

\begin{abstract}
This paper presents an omnidirectional aerial manipulation platform for robust and responsive interaction with unstructured environments, toward the goal of contact-based inspection. The fully actuated tilt-rotor aerial system is equipped with a rigidly mounted end-effector, and is able to exert a 6 degree of freedom force and torque, decoupling the system's translational and rotational dynamics, and enabling precise interaction with the environment while maintaining stability. An impedance controller with selective apparent inertia is formulated to permit compliance in certain degrees of freedom while achieving precise trajectory tracking and disturbance rejection in others. Experiments demonstrate disturbance rejection, push-and-slide interaction, and on-board state estimation with depth servoing to interact with local surfaces. The system is also validated as a tool for contact-based non-destructive testing of concrete infrastructure.
\end{abstract}

\IEEEpeerreviewmaketitle

\section{Introduction}
\label{sec:introduction}
The demand for aerial robotic workers for a wide range of applications has been steadily gaining the attention of research communities, industry, and the general public \cite{ruggiero2018aerial}.
As a compelling example, the status of aging concrete infrastructure is a growing concern due to the rising amount of required inspection, and a lack in capacity to meet the need by traditional means \cite{asce20172017}.
New technologies using \ac{NDT} contact sensors, such as potential mapping \cite{angst2018challenges}, permit detection of corrosion far earlier than visual assessment.
While aerial vehicles have already been embraced as a solution for efficient visual inspection of infrastructure \cite{chan2015towards}, contact-based \ac{NDT} still requires extensive human labor, road closure, and the use of large supporting inspection equipment.

Recent developments in omnidirectional aerial vehicles begin to bridge the gap, from using \acp{UAV} as efficient visual industrial inspection agents, to interacting with the structures they inspect.
The ability of omnidirectional aerial systems to exert a 6 \ac{DOF} force and torque allows for decoupling of the system's translational and rotational dynamics, enabling precise interaction with the environment while maintaining stability. However, making this solution a viable alternative to traditional inspection requires the design of an aerial system with on-board power and sensing, high force and torque capabilities in all directions, and precise and reliable interaction control in 6 \ac{DOF}.

We propose a novel omnidirectional aerial system: a hexarotor with actively tilting double propeller groups, resulting in high force and torque capabilities in all directions while maintaining efficient flight in horizontal hover. The configuration also permits omnidirectional orientation of the vehicle. The current work extends upon the system concept presented in \cite{bodie2018towards, kamel2018voliro}, with the addition of a rigidly attached end-effector for interaction. The simplicity of the system dynamic model as a single rigid body and the high degree of overactuation in propeller orientations allow for robust and responsive interaction and disturbance rejection. In this paper we present the system design of a fully actuated aerial manipulation platform capable of on-board computation, state estimation, power, and sensing, allowing for direct deployment in the field.

In particular, this paper presents the following contributions:
\begin{itemize}
    \item The system design of a novel omnidirectional tilt-rotor \ac{MAV} with a rigid manipulator arm.

    \item A 6 \ac{DOF} impedance control approach with selective apparent inertia for a fully-actuated flying system.

    \item Experimental validations showing precise interaction control, and demonstration of the system as a viable platform for contact-based \ac{NDT} of concrete infrastructure.
\end{itemize}

\begin{figure}
\centering
  \includegraphics[trim=0cm 0cm 0cm 0cm ,clip, width=0.48\textwidth]{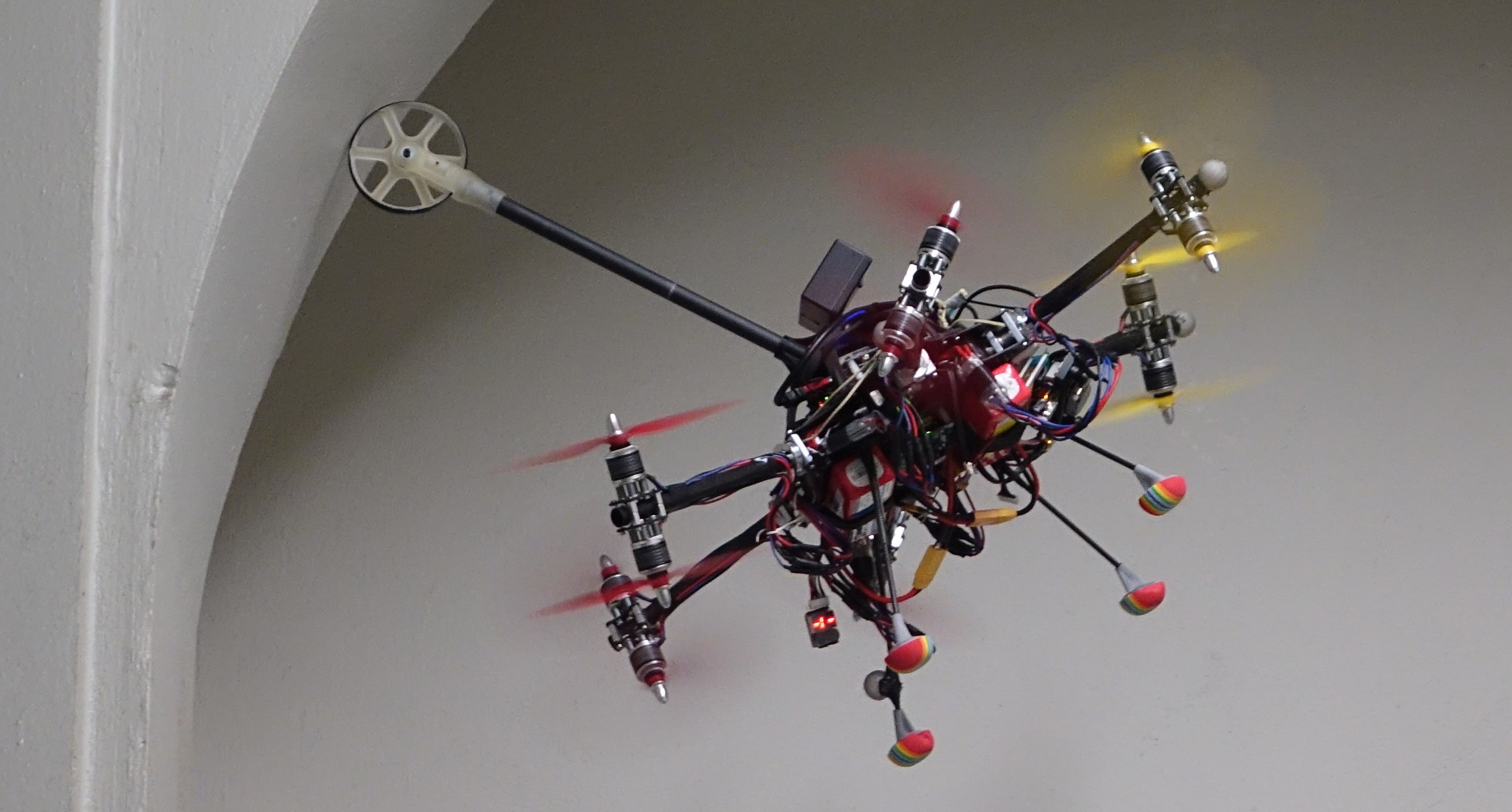}
\caption{Aerial manipulation platform in contact, rolling along a curved surface, with on-board state estimation.}
\label{fig:experiment_arch}
\end{figure}

\section{Related Work}
\label{sec:related_work}

Aerial interaction has been approached extensively in literature over the past decade with various systems \cite{ruggiero2018aerial}, addressing the problem of underactuation by adding \acp{DOF} in a manipulator arm, and handling the resulting dynamic complexity. While interaction control has been demonstrated on many underactuated systems equipped with manipulators \cite{alexis2016aerial,scholten2013interaction}, lateral force magnitudes and disturbance rejection capabilities are limited due to coupled rotational and translational dynamics.

In the past few years, fully actuated aerial manipulators have emerged to address these issues. Several platforms achieve full actuation by fixedly tilting the propeller groups of a traditional \ac{MAV} to generate internal forces that are optimized for interaction \cite{staub2017tele, staub2018towards, tognon2019truly, wopereis2018multimodal, ollero2018aeroarms}. These platforms are limited in roll and pitch, requiring additional \acp{DOF} in a manipulator arm. Full pose-omnidirectionality has been achieved in \cite{park2018odar} by an optimization of propeller orientations, and using a fixed end-effector for interaction. While the platform demonstrates a high down-force, lateral and upward force exertion are limited due to the internal forces needed for omnidirectionality. Recent publications have demonstrated push-and-slide interaction and disturbance rejection, as well as various industrial applications \cite{park2018odar, tognon2019truly, wopereis2018multimodal, ollero2018aeroarms}.
Our approach combines omnidirectionality with high force and torque capabilities to eliminate the need for an actuated manipulator arm. This results in simplified system dynamics without compromising disturbance rejection.

Control approaches for fully actuated systems with push-and-slide capabilities vary in implementation: One approach uses a cascaded \ac{PID} controller in free flight, switching to an angular rate stabilizing control when in contact \cite{wopereis2018multimodal}, another implements a dislocated \ac{PD} control law for an elastic jointed manipulator model, with integral action in all directions except along the tool's axis of contact \cite{tognon2019truly}, and a further approach uses a pose trajectory tracker in SE(3) in free flight, which switches to a hybrid pose/wrench control when in contact \cite{park2018odar}. Impedance control has been implemented on underactuated-base aerial manipulators \cite{lippiello2012exploiting, ruggiero2014impedance}, without omnidirectional force capabilities. The selective impedance control strategy presented in this paper seeks to advance this concept to address omnidirectional interaction with a single model-based controller. The proposed controller is used for all situations without reliance on transition handling, and using a planner for orientation and movement relative to the local surface normal.

A major challenge of porting such a system from a laboratory context to the real world is the need for good state estimation close to the environment. While many of the aforementioned systems operate in controlled environments that harness the precise and reliable capabilities of motion-capture state estimation, this approach is not viable in the field.
In the presented platform, an on-board \ac{VI} sensor enables state estimation in complex environments. Precise position and orientation relative to the environment are achieved with a 3D \ac{TOF} camera.

\section{System Description}
\label{sec:system_description}

\begin{figure}
\includegraphics[width=0.48\textwidth]{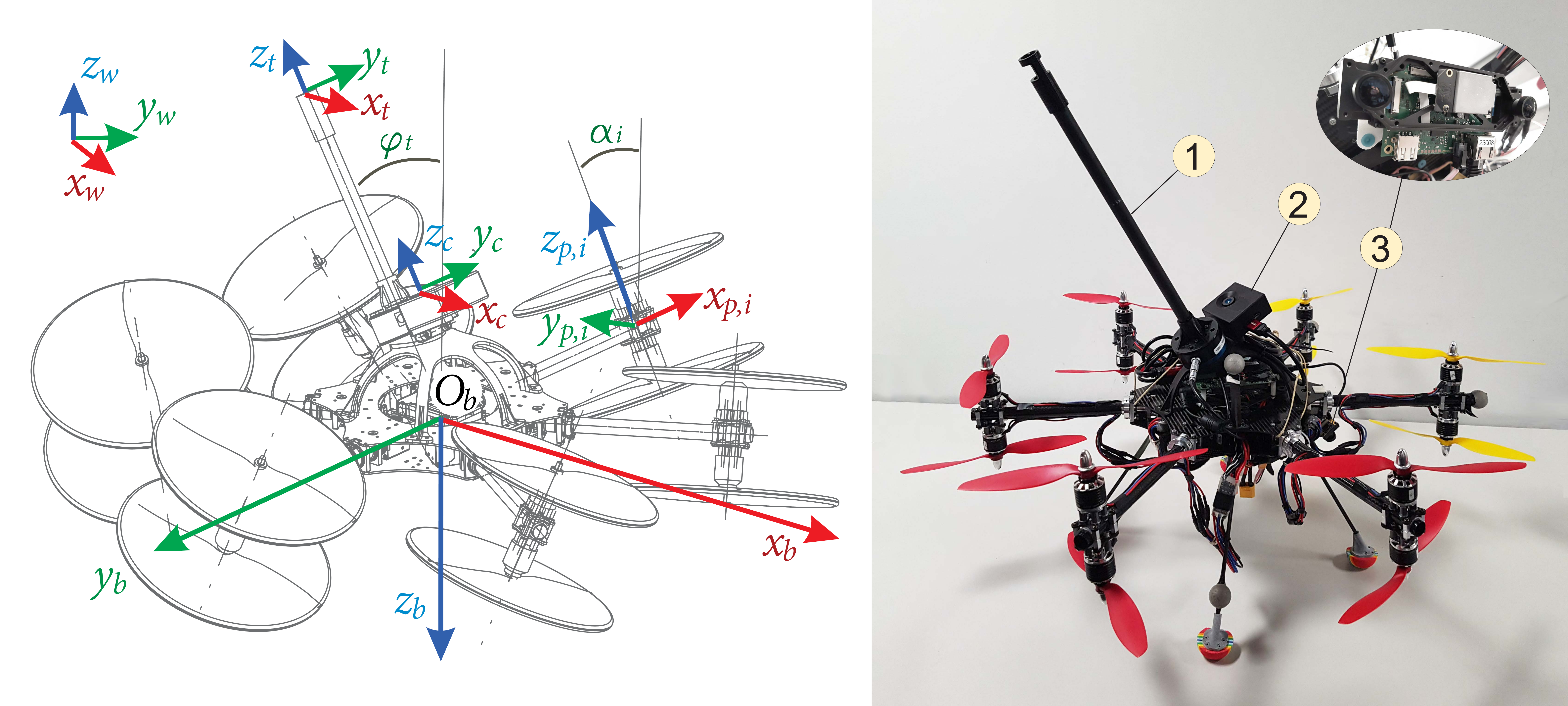}
 \caption{Coordinate frames, variables, and sensor components of the aerial manipulator: 1) a rigidly mounted arm at angle $\varphi_t^{\circ}$, 2) a \ac{TOF} camera used for local surface normal estimation, and 3) a \ac{VI} sensor for on-board state estimation.}
 \label{fig:description}
\end{figure}

The omnidirectional platform used in this work takes the form of a traditional hexarotor with equally spaced arms about the $z_b$-axis. Each propeller group is independently tilted ($\alpha_i$) by a dedicated servomotor located in the body, allowing for various rotor thrust combinations.
Double rotor groups provide additional thrust for a compact system. Symmetrically placed rotors balance rotational inertia about the tilt axis to reduce effort of the tilt motors.
Processing occurs on an on-board computer and flight controller. Lithium-polymer batteries are mounted to the system base for on-board power. Major system parameters are listed in \cref{tab:parameters}.

\begin{table}[htb!]
\vspace*{0.3cm}
\centering
\def\arraystretch{1.2}
\begin{tabular}{l c c}
\hline
\textbf{Parameter} & \textbf{Value} & \textbf{Units} \\
\hline
Total system mass & 4.75 & [\si{\kilogram}] \\
System diameter & 0.83 & [\si{\meter}] \\
Rotor group distance to $O_b$ & 0.3 & [\si{\meter}] \\
Contact arm pitch, $\varphi_t$ & 30 / 90 & [\si{\degree}] \\
Maximum thrust per rotor group & 20 & [\si{\newton}] \\
Number of rotor groups & 6 & \\
\hline
\end{tabular}
\caption{Main system parameters}
\label{tab:parameters}
\vspace*{0.3cm}
\end{table}

\begin{table}[htb!]
\centering
\def\arraystretch{1.2}
\begin{tabular}{l c c}
\hline
\textbf{Description} & \textbf{Subscript} \\
\hline
World frame & $_{W}$  \\
Body frame & $_{b}$ \\
Time-of-flight camera frame & $_{c}$ \\
Tool frame & $_{t}$ \\
\hline
\end{tabular}
\caption{Main coordinate frames}
\label{tab:frames}
\end{table}

A manipulator arm is rigidly mounted to the platform body, with a tool frame at the tip of the arm. The $z_t$-axis intersects the body origin, $O_b$, and lies on the $y_b$-plane. The arm orientation is fully defined by an angular declination of $\varphi_t$ from the negative $z_b$ axis. A \ac{TOF} camera is rigidly mounted near the base of the arm, its frame orientation aligned with the tool frame. A \ac{VI} sensor is oriented opposite to the tool to ensure reliable feature detection for state estimation during interaction.

\Cref{tab:frames} gives an overview of the frames used throughout this work.

\section{Control Framework}
\label{sec:control_framework}

\begin{figure*}[tp]
\centering
\includegraphics[trim=1cm 15.2cm 1cm 10cm, clip, width=0.85\textwidth]{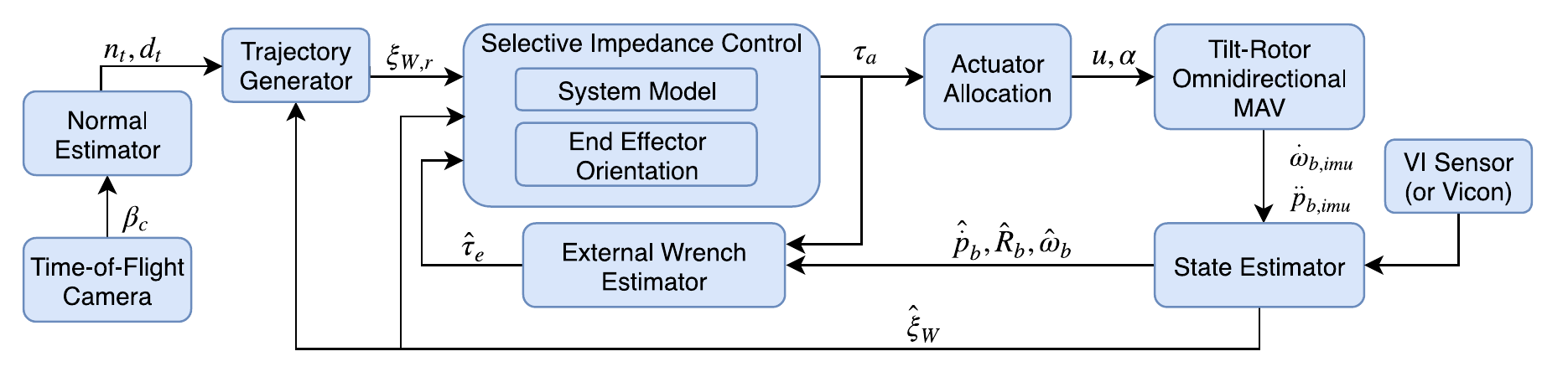}
 \caption{Block diagram of selective impedance control framework, with system state, $\xi = \{\bm{p},\dot{\bm{p}},\ddot{\bm{p}},\bm{R},\bm{\omega}, \dot{\bm{\omega}}\}$, and actuator commands, $\bm{u}$ and $\bm{\alpha}$. Surface normal, $\bm{n}_{t}$, and distance, $d_{t}$, are extracted from point cloud $\beta_{c}$ provided by the \ac{TOF} camera.}
 \label{fig:impedance}
\end{figure*}

A block diagram of the complete control framework is shown in \cref{fig:impedance}. This section describes the system model, selective impedance controller, external wrench estimator, and trajectory generator. Actuator allocation is handled according to \cite{bodie2018towards}, transforming a reference wrench into rotor speeds $\bm{u} \in \mathbb{R}^{12 \times 1}$ and tilt angles $\bm{\alpha} \in \mathbb{R}^{6 \times 1}$. Further details on the experimental setup are presented in \cref{sec:setup}.

\subsection{System Model}
To simplify the system model, we assume that the body is rigid, and that the body axes correspond with the principal axes of inertia. Thrust and drag torques are assumed proportional to squared rotor speeds $\omega_i$, and are instantaneously achievable without transients. We further assume that tilt motor dynamics are negligible compared to the whole system dynamics, and that airflow interference between propeller groups does not affect the net thrust and torque on the system.

The simplified system dynamics are derived in the Lagrangian form as
\begin{equation}
\bm{M} \dot{\bm{v}} + \bm{C} \bm{v} + \bm{g} = \bm{\tau}_{a} + \bm{\tau}_{e}
\label{eq:lagrangian}
\end{equation}

\noindent where $\bm{v}$ and $\dot{\bm{v}} \in \mathbb{R}^{6 \times 1}$ are stacked linear and angular velocity, and linear and angular acceleration of the origin, $\bm{M} \in \mathbb{R}^{6 \times 6}$ is the symmetric positive definite inertia matrix, $\bm{C} \in \mathbb{R}^{6 \times 6}$ contains the centrifugal and Coriolis terms, and $\bm{g} \in \mathbb{R}^{6 \times 1}$ is the gravity vector. The terms $\bm{\tau}_{a}$ and $\bm{\tau}_{e} \in \mathbb{R}^{6 \times 1}$ are stacked force and torque vectors exerted on the system respectively by rotor actuation and external sources (e.g. contact or wind disturbances).

\subsection{External Wrench Estimation}
In order to account for the influence of contact forces, we employ an external wrench estimator using a generalized momentum approach. Our implementation follows the method described in \cite{ruggiero2014impedance}, and is expressed as follows:

\begin{equation}
\hat{\bm{\tau}}_{e} = \bm{K}_I\left(\bm{M}\bm{v}- \int\left(\bm{\tau}_{a} - \bm{C}\bm{v} - \bm{g} + \hat{\bm{\tau}}_{e}\right)dt\right)
\label{eq:wrench_estimation}
\end{equation}

The positive-definite diagonal observer matrix $\bm{K}_I \in \mathbb{R}^{6 \times 6}$ acts as an estimator gain. Differentiating \eqref{eq:wrench_estimation}, a first-order low-pass filtered estimate $\hat{\bm{\tau}}_{e}$ of the external wrench $\bm{\tau}_{e}$ is obtained:

\begin{equation}
\dot{\hat{\bm{\tau}}}_{e} = \bm{K}_I(\bm{\tau}_{e} - \hat{\bm{\tau}}_{e})
\label{eq:firstOrderLowPass}
\end{equation}

Note that \eqref{eq:wrench_estimation} allows estimation of external forces and torques without the use of acceleration measurements, but instead it only requires linear and angular velocity estimates.

\subsection{Interaction Control: Selective Impedance}
Impedance control indirectly regulates a wrench exerted by the system on its environment, without the drawbacks of contact detection and controller switching that accompany direct force control. The same controller can account for interaction, as well as for stable flight in free space. We can take advantage of the system's full actuation to implement an impedance controller with selective apparent inertia, to reject disturbances in some directions while exhibiting compliant behavior in others.
Our implementation is based on the method described in \cite{siciliano2010robotics}. We take the simplified dynamics of the system as \eqref{eq:lagrangian} and choose the desired closed loop dynamics of the system to be
\begin{equation}
\bm{M}_v\dot{\bm{v}} + \bm{D}_v\widetilde{\bm{v}} + \bm{K}_v\widetilde{\bm{x}} = \bm{\tau}_{e},
\label{eq:desired_dynamics}
\end{equation}
where $\bm{M}_v, \bm{D}_v,$ and $\bm{K}_v \in \mathbb{R}^{6 \times 6}$ are positive-definite matrices representing the desired apparent inertia, desired damping, and desired stiffness of the system. Position and velocity errors are represented by $\widetilde{\bm{x}}$ and $\widetilde{\bm{v}} \in \mathbb{R}^{6 \times 1}$, respectively.
We can then derive the applied control wrench by substituting $\dot{\bm{v}}$ from \eqref{eq:desired_dynamics} into \eqref{eq:lagrangian} as follows:
\begin{equation}
\bm{\tau}_{a} = (\bm{M}\bm{M}_v^{-1} - \mathbb{I}_3)\hat{\bm{\tau}}_{e} - \bm{M}\bm{M}_v^{-1}(\bm{D}_v\widetilde{\bm{v}} + \bm{K}_v\widetilde{\bm{x}}) + \bm{C}\bm{v}
+ \bm{g}
\label{eq:impedance1}
\end{equation}

Since the stiffness and damping properties of interaction depend highly on the apparent inertia, we first normalize these matrices with respect to the system mass as $\Tilde{\bm{M}}_v = \bm{M}^{-1}\bm{M}_v$, then express stiffness and damping as $\Tilde{\bm{D}}_v = \Tilde{\bm{M}}_v^{-1}\bm{D}_v$ and $\Tilde{\bm{K}}_v = \Tilde{\bm{M}}_v^{-1}\bm{K}_v$. In addition, the selective impedance can be rotated into a desired frame, in this case the fixed end-effector frame, using $\bm{R} = \text{blockdiag}\{\bm{R}_{bt}, \bm{R}_{bt}\}$, where $\bm{R}_{bt}$ is a rotation matrix expressing the orientation of the tool frame in the body-fixed frame. We then rewrite \eqref{eq:impedance1} as
\begin{equation}
\bm{\tau}_{a} = (\bm{R}^\top \Tilde{\bm{M}}_v^{-1} \bm{R} - \mathbb{I}_3)\hat{\bm{\tau}}_{e} - \Tilde{\bm{D}}_v\widetilde{\bm{v}} - \Tilde{\bm{K}}_v\widetilde{\bm{x}} + \bm{C}\bm{v} + \bm{g}.
\label{eq:impedance2}
\end{equation}

Integration of a rigidly attached end-effector to the system simplifies the problem of selective stiffness in impedance control. The apparent mass is lowered along the $z_t$-axis. Orienting the $z_t$-axis normal to the desired contact surface is then sufficient to ensure compliance in the contact direction and stiff behavior in the orthogonal plane, and a stiff response to error in orientation.

\subsection{Surface Normal and Distance Estimation}
\begin{figure}
\centering
\subfloat{
  \includegraphics[clip,width=0.48\textwidth]{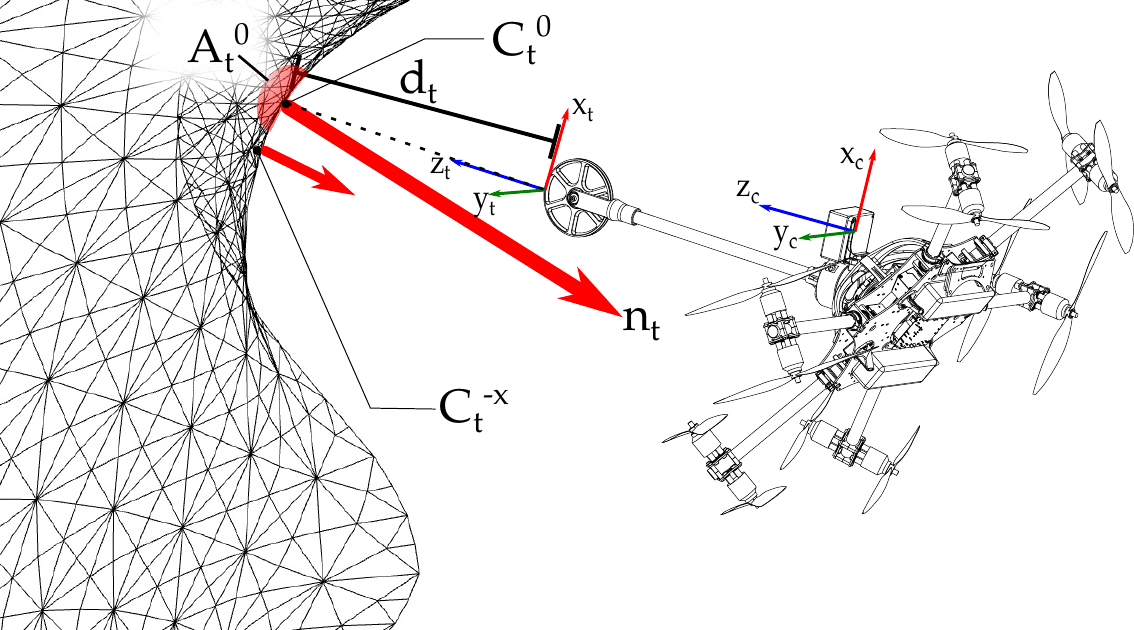}}
\caption{Normal estimation and path planning: $C_{t}^{0}$ is the contact point defined by the intersection of the $z_{t}$-axis and the observed surface. The red arrow $n_{t}$ depicts the estimated surface normal, $d_{t}$ the distance to the contact point, and $A^{0}_{t}$ is the set of 3D points used to estimate normal and distance. $C^{-x}_{t}$ represents a set point for moving along the $-x_{t}$-axis.}
\label{fig:planning}
\end{figure}

For surface normal and distance estimation, the desired contact point is defined as the intersection of an observed surface and the $z_t$-axis ($C^{0}_{t}$ in \cref{fig:planning}).
In order to estimate the distance to this point and to compute the local surface normal, a point cloud is obtained from the \ac{TOF} camera and all points within a certain distance to the $z_t$-axis are selected. This subset of 3D points is subsequently called $A^{0}_{t}$.
The 3D contact point location is obtained by an unweighted average of all points in $A^{0}_{t}$.
The local surface normal is obtained by fitting a plane to the set of points $A^{0}_{t}$ in a least-squares sense.
If the resulting vector $n_{t}$ points away from the camera ($n_{t} \cdot z_{t} > 0$), the direction is flipped to ensure consistency.

\subsection{Path Planning in 6 \ac{DOF} on a Manifold}
The \ac{MAV} is maneuvered to a position relatively close to the surface while maintaining a level orientation. As soon as a stable normal and distance estimate are obtained, the target set point in the tool frame is calculated. The target set point is placed a certain distance behind the estimated contact point. This ensures reliable contact thanks to the impedance controller, even in the presence of small fluctuations in state or distance estimation. The target orientation of the \ac{MAV} is chosen such that the $z_{t}$-axis aligns with the estimated surface normal. Rotation is fully defined by requiring the body $y_{b}$-axis to be aligned with the ground plane, defined by $y_{W}$/$x_{W}$, constraining rotation about the $z_{t}$-axis. A smooth trajectory is generated such that the tool is driven to the estimated target pose.

Translation along the surface is achieved by moving the set point tangent to the currently estimated surface in a desired direction (e.g. $C_{t}^{-x}$ in \cref{fig:planning}). The set point as well as the surface estimate and generated trajectory are updated at a rate of \SI{5}{\hertz}. Effectively, this corresponds to a pure pursuit path tracking that always maintains contact and orientation with respect to the surface.

\section{Experiments}
\label{sec:experiments}
\subsection{Experimental Setup}
\label{sec:setup}

\begin{table}
\centering
\def\arraystretch{1.2}
\begin{threeparttable}
\begin{tabular}{l c c c c c c c}
\hline
\textbf{Experiment} & \textbf{$\varphi_t$} [$^{\circ}$] &
\textbf{$m^*_x$} & \textbf{$m^*_y$} & \textbf{$m^*_z$} &
\textbf{$I^*_{xx}$} & \textbf{$I^*_{yy}$} & \textbf{$I^*_{zz}$} \\
\hline
Rope pull 1     & 90 & 0.25 & 0.25 & 1.0 & 1.0 & 1.0 & 1.0 \\
Rope pull 2     & 90 & 0.1 & 0.1 & 1.0 & 1.0 & 1.0 & 1.0 \\
Rope pull 3     & 90 & 5.0 & 5.0 & 1.0 & 1.0 & 1.0 & 1.0 \\
Rope pull 4     & 90 & 5.0 & 5.0 & 5.0 & 5.0 & 5.0 & 0.25 \\
Rope pull 5     & 90 & 5.0 & 5.0 & 5.0 & 5.0 & 5.0 & 5.0 \\
Push and slide   & 90 & 5.0 & 5.0 & 0.25 & 5.0 & 5.0 & 5.0 \\
ToF servoing    & 30 & 5.0 & 5.0 & 0.5 & 5.0 & 5.0 & 5.0 \\
Contact NDT     & 90 & 5.0 & 5.0 & 0.25 & 5.0 & 5.0 & 5.0 \\
Force eval 1     & 90 & 5.0 & 5.0 & 0.25 & 5.0 & 5.0 & 5.0 \\
Force eval 2     & 90 & 5.0 & 5.0 & 2.0 & 5.0 & 5.0 & 5.0 \\
\hline
\end{tabular}
\caption{Tool and impedance parameters for experiments}
\label{tab:impedance_params}
\vspace*{0.5cm}
\begin{tablenotes}
  \item[] $^*$Inertial parameters are multipliers of the system inertia.
\end{tablenotes}
\end{threeparttable}

\vspace*{-\baselineskip}
\end{table}

Experiments are conducted using the tilt-rotor aerial manipulation platform presented in \cref{sec:system_description} with characteristics shown in \cref{tab:parameters}, and specific test parameters shown in \cref{tab:impedance_params}. All diagonal values of $\bm{K}_I$ for the wrench estimator are set to unity. A safety tether is connected loosely to the robot, and minimally affects results. A supplementary video\footnote{\href{https://youtu.be/-RCQmaKvsL0}{https://youtu.be/-RCQmaKvsL0}} is also available for reference.

Several experiments fuse Vicon motion capture data sampled at \SI{100}{\hertz} with on-board \ac{IMU} measurements sampled at \SI{250}{\hertz} for state estimation (henceforth referred to as ``external state estimation").
For viability in industrial environments, however, on-board sensing is preferred. In \cref{sec:experiment_tof}, only on-board sensor data is processed.
The \ac{VI} state estimation framework Rovio \cite{bloesch2015robust} is used to determine position and attitude (henceforth referred to as ``on-board state estimation") in combination with a \ac{TOF} camera that estimates distance and orientation w.r.t. the contact surface.

The experiments in this section are designed to demonstrate the following:

\begin{itemize}
\item System response to an external wrench with selective impedance control, demonstrating disturbance rejection (\cref{sec:experiment_rope}).
\item Repeatable tracking of position and orientation when interacting with a planar surface, while rejecting disturbances due to surface friction (\cref{sec:experiment_whiteboard}).
\item Ability to interact with complex surfaces using purely on-board sensing (\cref{sec:experiment_tof}).
\item Viability as an infrastructure \ac{NDT} contact testing tool (\cref{sec:experiment_ndt}).
\end{itemize}

\subsection{Rope Pull Disturbance in Free Flight}
\label{sec:experiment_rope}

\begin{figure}
\centering
\subfloat{
  \includegraphics[trim=0cm 0cm 0cm 0cm, width=0.472\textwidth]{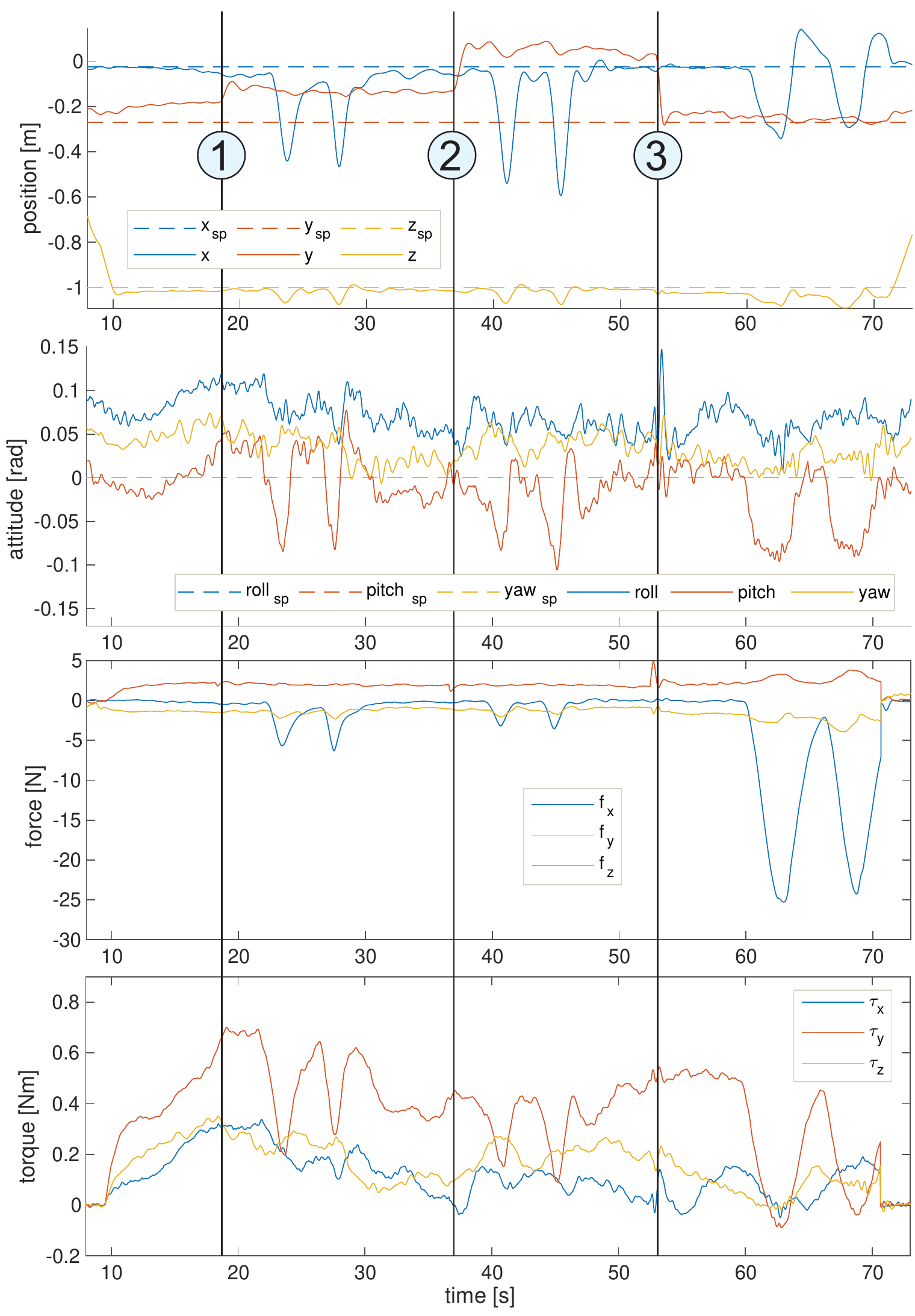}}
\caption{Pose tracking and wrench estimation for rope pulling at the tool tip along the $z_t$-axis in free flight.}
\label{fig:experiment_rope_1}
\end{figure}

\begin{figure}
\centering
\subfloat{
  \includegraphics[trim=0cm 0cm 0cm 0cm, width=0.48\textwidth]{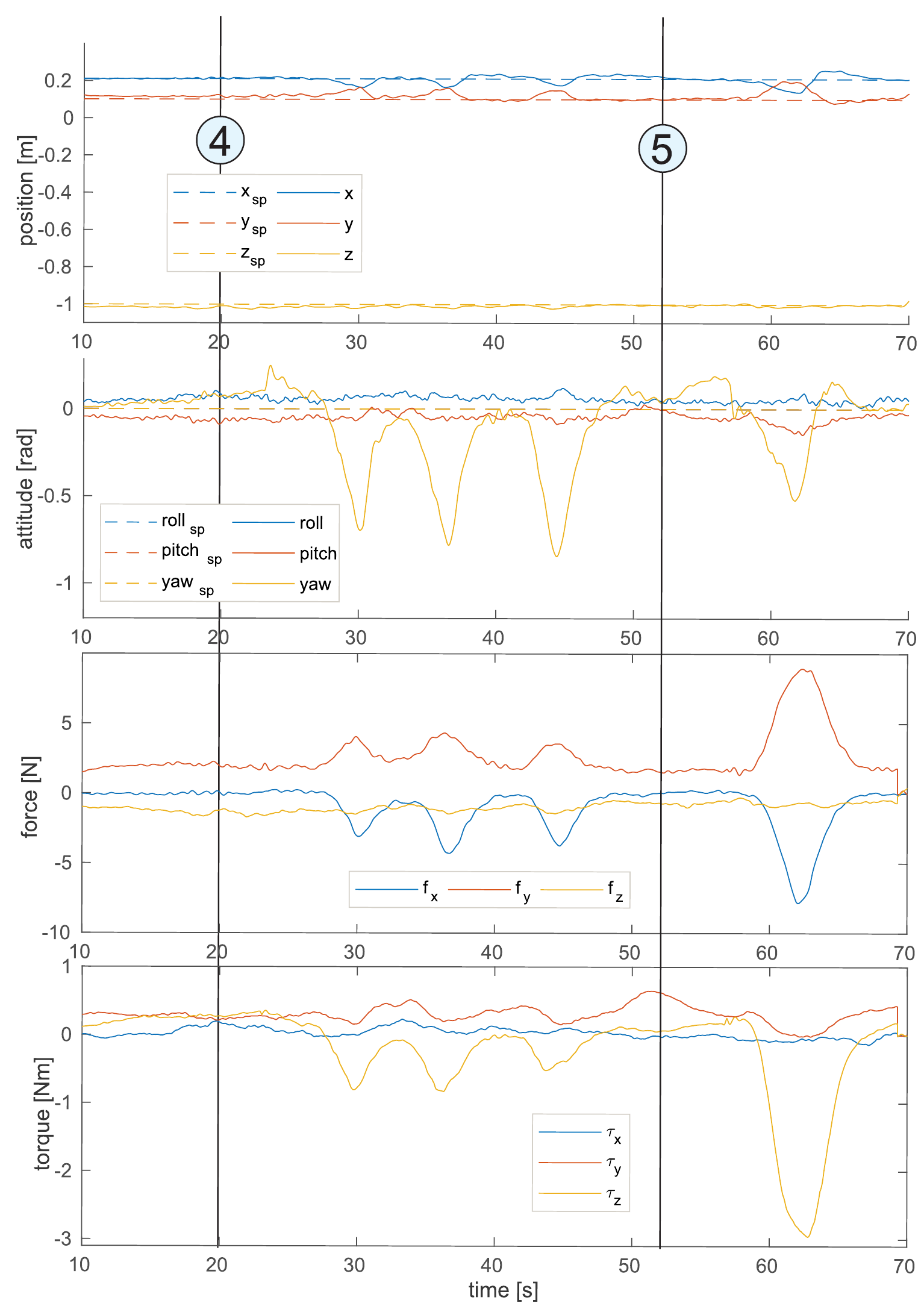}}
\caption{Pose tracking and wrench estimation for rope pulling at the tool tip orthogonal to the $z_t$-axis in free flight.}
\label{fig:experiment_rope_2}
\end{figure}

In this experiment, we evaluate the behaviour of the system with different selective apparent inertia values, demonstrating the ability to reject large disturbances in certain directions.
The system is commanded to hold a reference pose \SI{1}{\meter} above the ground in free flight. A cord is tied to the tool tip of the rigid manipulator arm, which is aligned with the $x_b$-axis. The other end of the cord is pulled manually to generate an external wrench.
The test is performed indoors with external state estimation.
In tests 1 through 3, with results shown in \cref{fig:experiment_rope_1},
two pulls of the rope are made for each set of apparent inertia parameters, approximately along the negative $x_b$-axis.
In tests 4 and 5, with results shown in \cref{fig:experiment_rope_2},
a pull force is applied horizontally perpendicular to the fixed arm axis to generate a torque about the $z_b$-axis.

Tests 1 and 2 show similar results: a compliant response to a disturbance force in the direction of pull. Apparent mass values in $x_b$ and $y_b$ are lower than the actual system mass, meaning that force disturbances in these directions will be tracked in the controller, while the \ac{PD} component simultaneously tracks the reference trajectory.
Results in test 2 show a larger movement in response to a smaller applied force in low impedance directions, relative to test 1. The remaining degrees of freedom have high apparent inertia values, actively rejecting detected disturbances to track the reference trajectory.
In test 3, apparent mass in $x_b$ and $y_b$ are set to 5 times the system mass and inertia. Results show positional movement of less than \SI{0.3}{\meter} under a lateral disturbance force of \SI{25}{\newton}, demonstrating an ability to actively reject large force disturbances.

Tests 4 and 5 compare the response to a torque disturbance about the $z_b$-axis with apparent rotational inertia less than, and greater than the system inertia. In test 4, three pulls are made, targeting rotation about the $z_b$-axis, and rotational compliance is clearly shown in yaw in the attitude tracking plot. While some additional torques are generated around the remaining axes, these are actively rejected by the controller.
In test 5, the system counteracts a rotational torque of \SI{3}{\newton\meter} magnitude, reducing the yaw deviation to \SI{0.5}{\radian}.
High apparent mass in all directions successfully rejects forces up to \SI{8}{\newton} with a translational deviation of less than \SI{0.1}{\meter}.
These results motivate impedance parameters chosen for the remaining experiments.

\subsection{Push-and-Slide Along a Flat Surface}
\label{sec:experiment_whiteboard}

In this experiment, we evaluate the ability of the system to maintain a normal orientation to a whiteboard, rejecting disturbances from friction forces when interacting, as well as the ability to accurately and repeatably draw a defined pattern on the surface. The whiteboard is positioned in a known location, and a trajectory traces a drawing with the tool point \SI{10}{\cm} behind the surface of the whiteboard. The end-effector is equipped with a standard whiteboard marker, with no additional compliance. Apparent mass is set high in all directions, except for the $z_t$-axis, where it is set to be compliant. Refer to \cref{tab:impedance_params} for apparent impedance parameters. The experiment uses external state estimation, and takes place in an indoor environment.

\begin{figure}
\centering
\subfloat{
\includegraphics[width=0.45\textwidth]{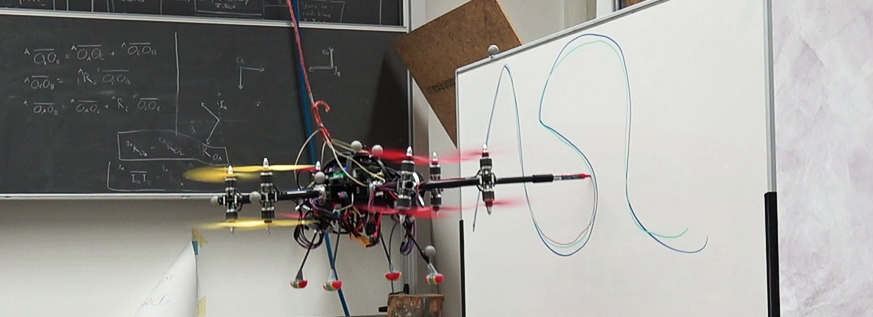}}
\hspace{0mm}
\subfloat{
  \includegraphics[trim=0cm 0cm 0cm 0cm, width=0.48\textwidth]{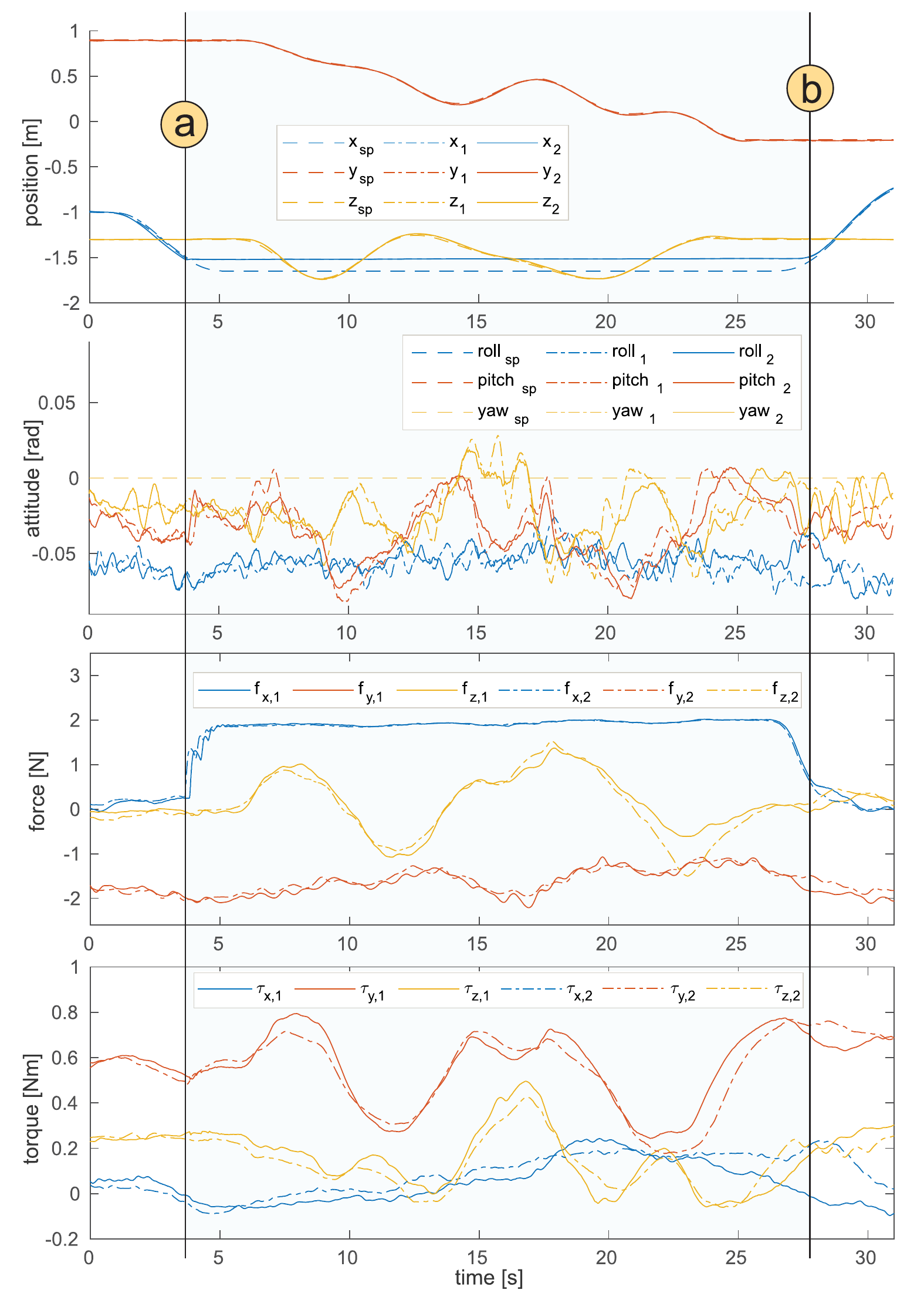}}
\caption{Pose tracking and wrench estimation for push-and-slide experiments. A shape is drawn on a whiteboard aligned with the $x$-plane in two separate trials (subscripts 1 and 2) with Vicon state estimation. At time (a), the system contacts the wall, and at (b), resumes free flight.}
\label{fig:whiteboard}
\end{figure}

Tracking results for position and orientation in the top two plots of \cref{fig:whiteboard} show ground truth measurements from the motion capture system of two trials drawing the same shape on a whiteboard, compared to the reference trajectory, marked with subscript $sp$, for set point. In the time interval between (a) and (b), the tool is in contact with the whiteboard, maintaining a consistent force while completing a trajectory.

Without any change in the controller, the system is able to handle transitions in and out of contact with good stability, and without noticable tracking error on the surface plane.
The system demonstrates a good ability to reject torque and lateral force disturbances caused by surface friction while maintaining a consistent contact force against the wall, as shown in the lower two plots of \cref{fig:whiteboard}.

Offsets in $y_b$-force and $y_b$- and $z_b$-torque in free flight---as well as a small bias in the attitude tracking of the system---are the result of an unaccounted-for offset of the system's center of mass. This result demonstrates that the proposed impedance control can compensate well for model errors, maintaining attitude error within \SI{0.07}{\radian}.

\subsection{Depth Servoing for Contact with Unknown Surfaces}
\label{sec:experiment_tof}

\begin{figure}
\centering
\subfloat{
  \includegraphics[width=0.45\textwidth]{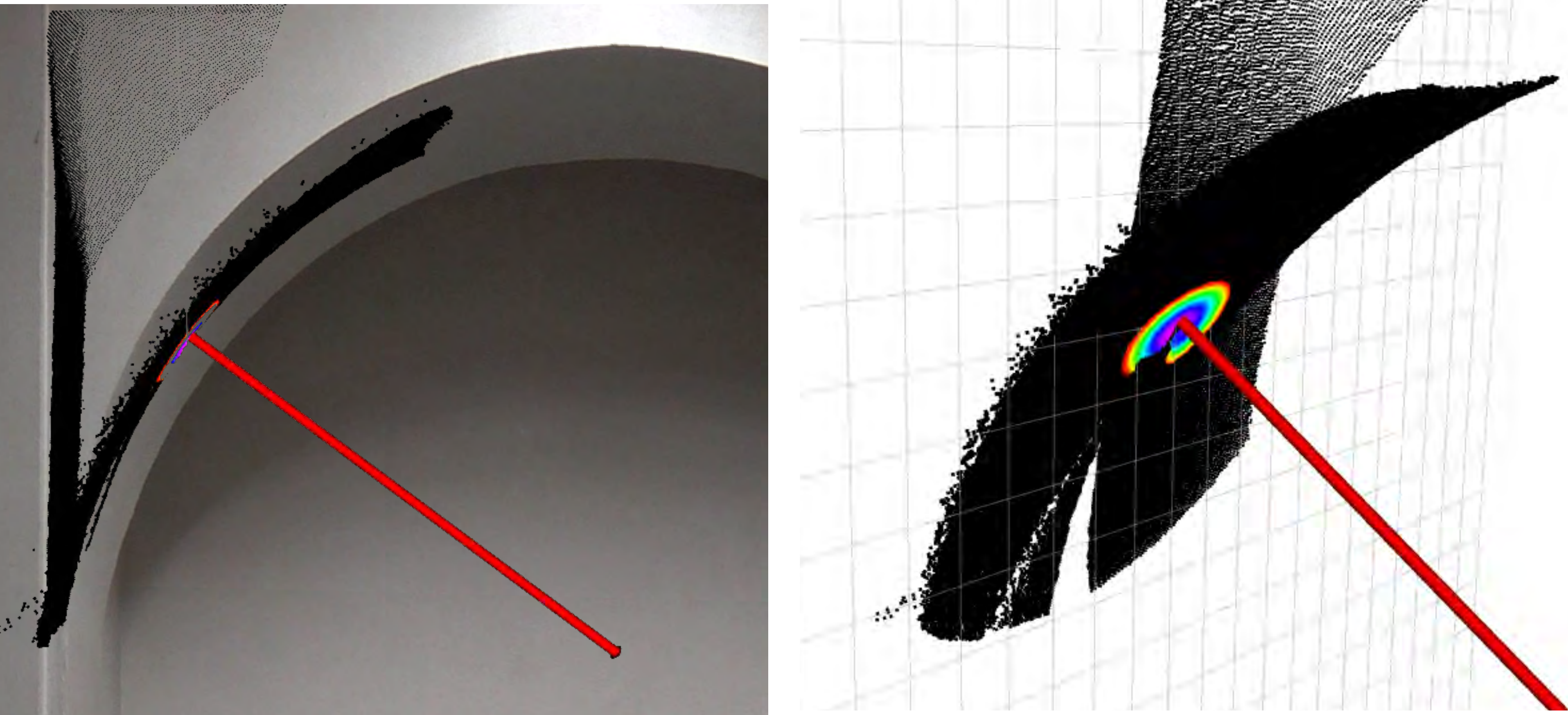}
  }
\caption{Visualization of a point cloud obtained from the ToF camera. Left image shows the overlay of the point cloud (black) on a photograph of the test site. The red arrow corresponds to the estimated normal. The same scene is depicted on the right image. The colored points represent $A^{0}_{t}$ and are used for the normal estimate.}
\label{fig:normal_estimation}
\end{figure}

This experiment evaluates the ability to use depth servoing to maintain orientation to a local surface normal in unknown environments, using a state estimation strategy that is directly deployable in the field.
The system is manually positioned to face a start point, then autonomously approaches the structure while orienting the $z_t$-axis to align with the locally observed surface normal (see \cref{fig:normal_estimation}). Contact is made, a line is traced, and the system leaves the structure.
State estimation is achieved using on-board sensing only, and tests are performed without any prior information such as maps or structural information.
The test environment is the ceiling vault of a staircase landing that is in daily use (see \cref{fig:experiment_arch}). The maximum altitude of contact is approximately \SI{4}{\meter} above the ground.

\begin{figure}
\centering
\subfloat{
  \includegraphics[width=0.45\textwidth]{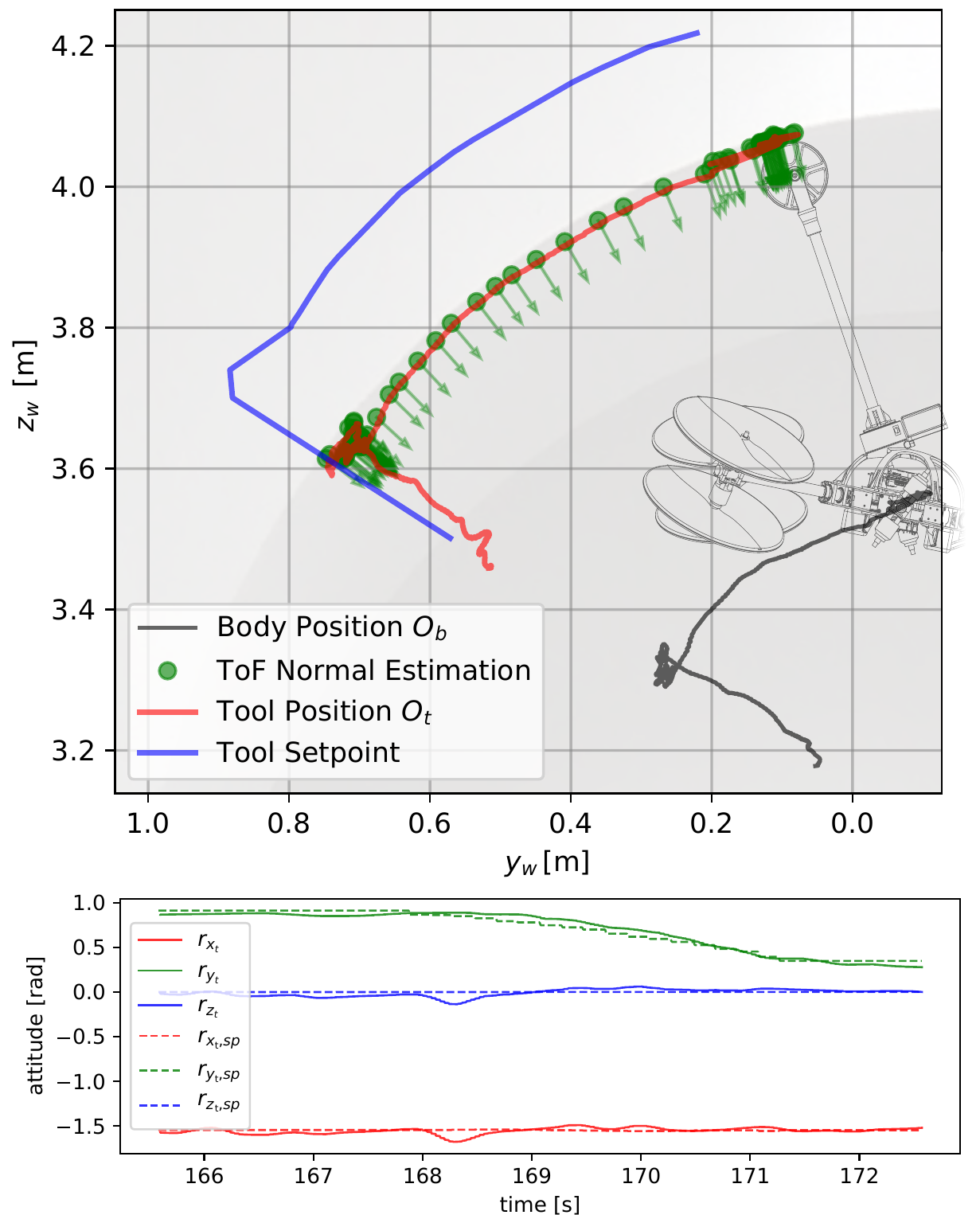}
 }
\caption{Top: Side view of pose tracking and normal estimates on a ceiling vault, orienting to surface using a \ac{TOF} sensor and on-board state estimation.
Bottom: Attitude tracking and set point in the tool frame based on the normal estimate. Translation along the surface occurs between \SI{168}{\second} and \SI{171}{\second}.}
\label{fig:experiment_tof}
\end{figure}

As visualized in \cref{fig:experiment_tof}, the estimated surface normals and distances are smooth and consistent which subsequently results in a clean set point trajectory and reliable contact.
At the initial contact point, the on-board state estimator drifts several centimeters. As the surface normal and distance estimates are drift-free and planning is updated frequently, our system is robust against such drifts.
Overall, the tool traverses a distance of \SI{0.53}{\meter} on the surface during contact, with an average velocity of \SI{0.17}{\metre\per\second} .

The bottom plot in \cref{fig:experiment_tof} shows the set point and measured attitude of the tool frame. During the time frame shown, attitude is set based on the estimated surface normal. The change of attitude during translation along the surface only affects rotation about the $y_{t}$-axis.

\subsection{Tests on Concrete Block with NDT Contact Sensor}
\label{sec:experiment_ndt}

In this experiment, we evaluate the ability of the system to maintain the positional accuracy and force required for useful measurements with a \ac{NDT} contact sensor on reinforced concrete structures.
The test is performed with external state estimation.

The rigid arm is equipped with a \ac{NDT} contact sensor that measures both the electrical potential difference between a saturated \ac{CSE} and the embedded steel, and the electrical resistance between the sensor on the concrete surface and the steel reinforcement.
Electrical potential and resistance results can be used as an indicator for the corrosion state of the steel \cite{bertolini2013corrosion}. A cable is connected to the reinforcement in the concrete structure, and is routed to the flying system via a physical tether to perform the measurements.
The concrete specimen used for this experiment has a known corrosion spot at a certain location and a constant cover depth within the block. The corrosion state can therefore be evaluated against this information.
The concrete block is positioned at a known location, and a trajectory is defined to contact 9 points at \SI{5}{\cm} intervals along the surface, with the contact point set \SI{10}{\cm} behind the surface of the wall to generate sufficient contact force for meaningful measurements.

\begin{figure}
\centering
\subfloat{
  \includegraphics[width=0.43\textwidth]{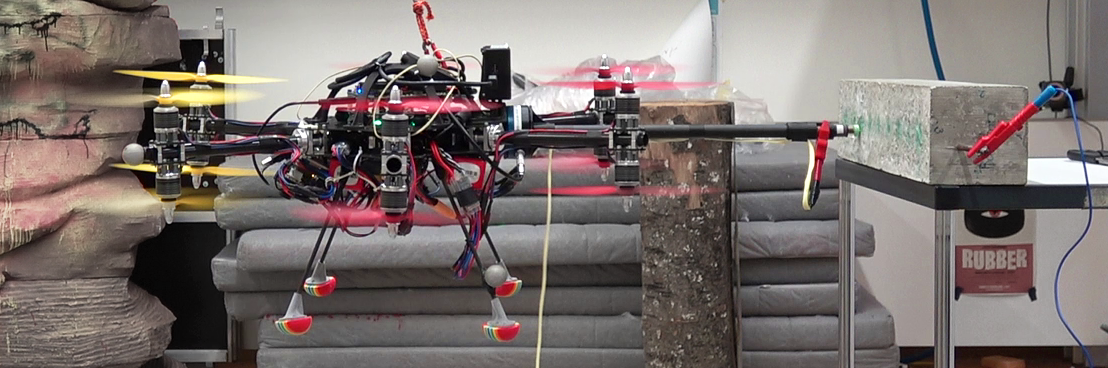}}
\hspace{0mm}
\subfloat{
  \includegraphics[trim=3.4cm 11cm 3.2cm 6cm, clip, width=0.5\textwidth]{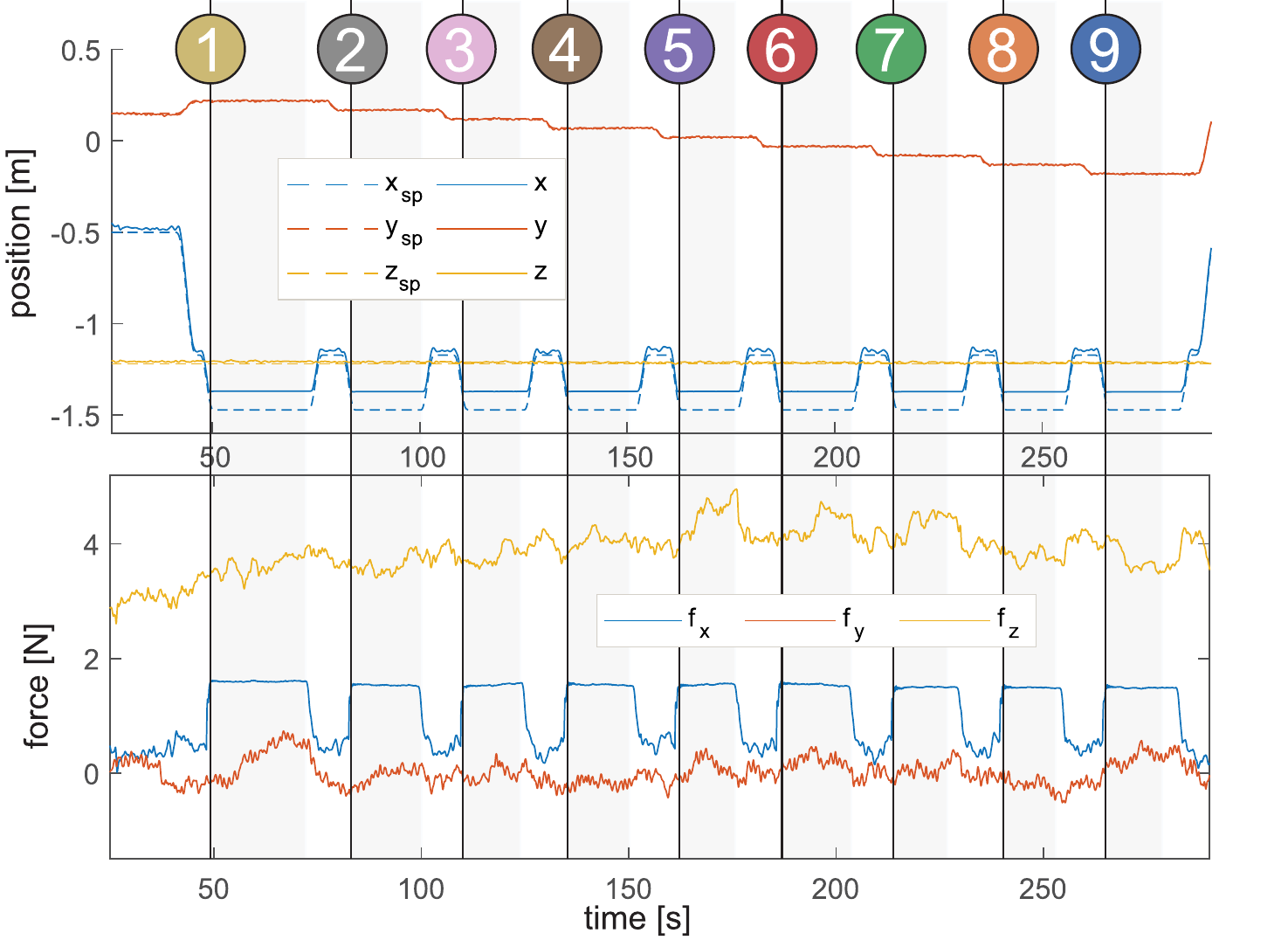}}
\caption{Position tracking and force estimation for tests with a \ac{NDT} contact sensor on a concrete block, aligned with the $x_b$-plane. Nine measurements are taken at \SI{5}{\cm} intervals with external state estimation, grey areas indicate contact.}
\label{fig:experiment_ndt}
\end{figure}

Tracking results in the top plot of \cref{fig:experiment_ndt} show precise trajectory tracking along all translational axes, except during shaded contact regions, where the $x_W$ position is blocked by the concrete specimen. A low apparent inertia in the $z_t$-direction allows for compliant behaviour of the system. The second plot shows forces that arise in the direction of the concrete surface, achieving a value of \SI{1.8}{\newton} in the contact phase. A constant offset in $z_b$-force can be seen, and is attributed to an error in the system mass and the center of mass offset estimates. Tracking results demonstrate that the controller is robust to this model error in directions with high apparent inertia.

\begin{figure*}[tb]
\centering
\subfloat{
  \includegraphics[width=0.88\textwidth]{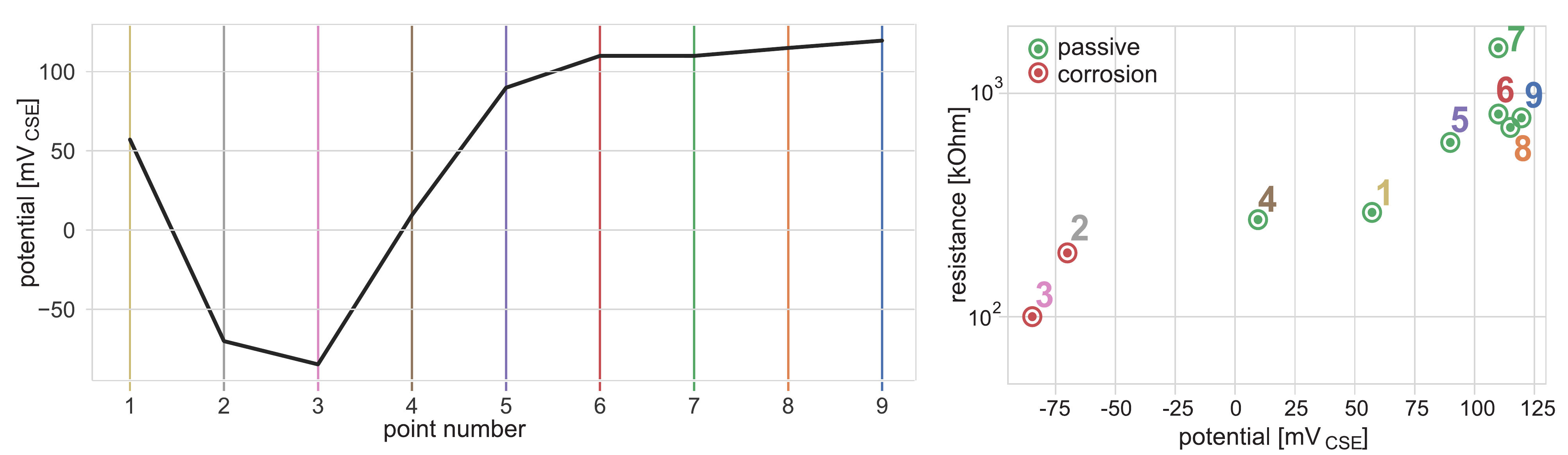}}
\caption{(left) Half-cell potential mapping with 9 points on a concrete specimen reinforced with carbon steel and \\ (right) corrosion state analysis for each point based on resistance and potential measured by an \ac{NDT} contact sensor.}
\label{fig:measurements_ndt}
\end{figure*}

 Data collected from the \ac{NDT} sensor are shown in \cref{fig:measurements_ndt}, where corrosion implications are deduced according to \cite{jaggi2005chlorinduzierte}, and correspond with corrosion at contact points 2 and 3.

\subsection{Evaluation of Force and State Estimators}
\label{sec:wrench_gt}

Additional tests were performed to evaluate on-board state estimation as a viable alternative to a motion capture system, and to determine the accuracy of estimated force measurements. On-board state estimation is used in flight, with data from a Vicon motion capture system as ground truth. Estimated force is computed using on-board state estimation as an input to the generalized momentum approach. Force ground truth data was collected from a 6-axis Rokubi 210 force sensor with its surface aligned with the $z_{W}$-plane, rigidly mounted to a wall. The force sensor records measurements at \SI{800}{\hertz}, with a resolution of \SI{0.1}{\newton}. The system follows a trajectory along the $x_{W}$-axis which goes in and out of contact with the force sensor. Apparent mass in the direction of contact is changed between contact events.

\begin{figure}
\centering
\includegraphics[trim=0cm 0cm 0cm 0cm, clip, width=0.48\textwidth]{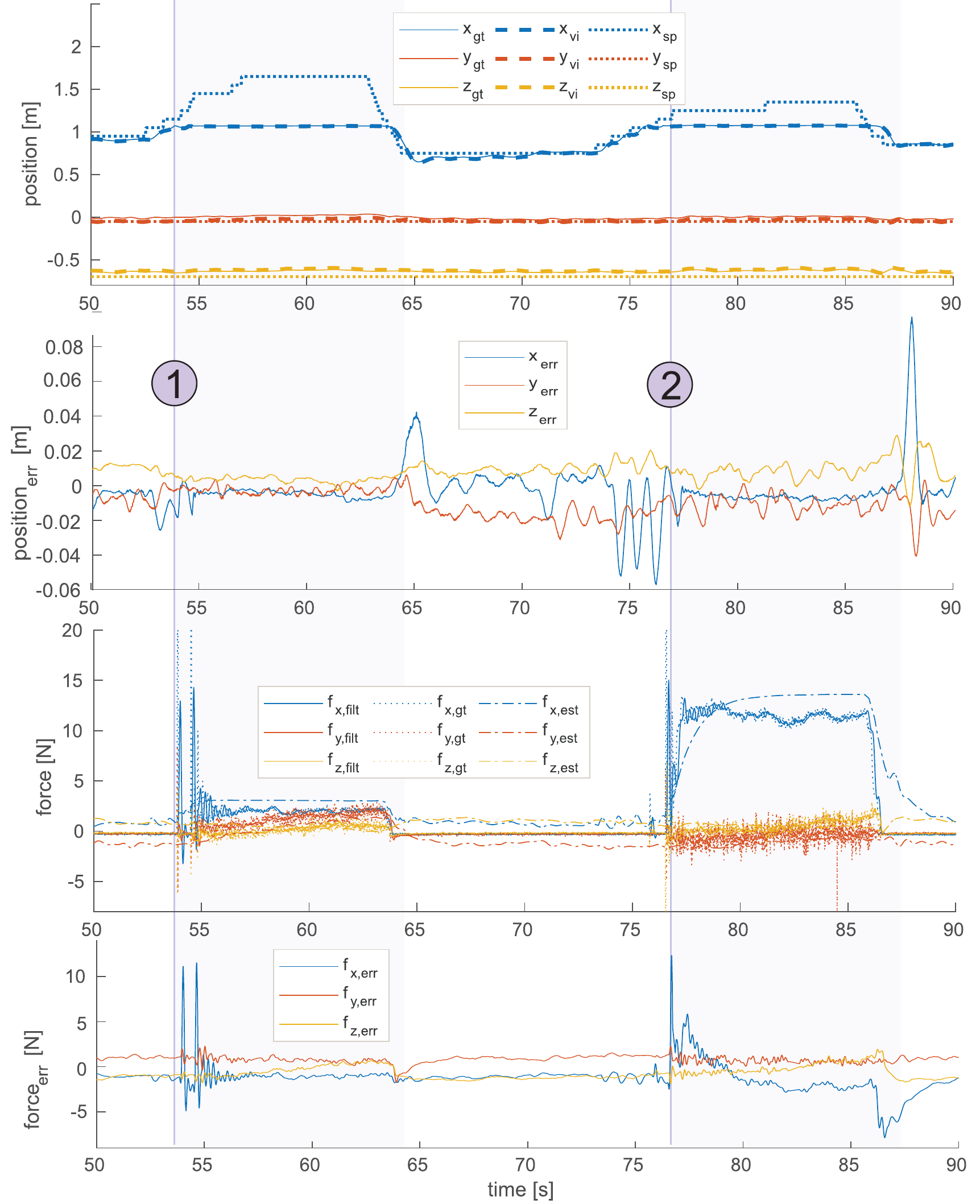}
\caption{On-board state estimation is compared with Vicon ground truth, tracking a set point to contact the force sensor. At points (1) and (2) the tool contacts the force sensor for the duration of the shaded region. Force estimates are plotted against raw and filtered force sensor data.}
\label{fig:ground_truth}
\end{figure}

The first plot in \cref{fig:ground_truth} compares positional on-board state estimation to ground truth, as the system tracks a trajectory along the $x_{W}$-axis. At contact points (1) and (2), movement along $x$ is blocked by the force sensor. The system remains stable under impedance control despite a set point offset of \SI{0.6}{\meter}. Error in on-board state estimation is presented in the second plot, showing error growth when the system is in motion versus small error in quasi-static hovering. \Ac{RMSE} of all 6 positional and rotational measurements are presented in \cref{tab:gt_error}. Roll and pitch errors are higher due to a bias, which we can attributed to a center of mass offset. Results show that on-board state estimation can adequately replace an external motion capture system for interactive flights of short duration.

The lower plots in \cref{fig:ground_truth} compare momentum based force estimation to ground truth. Due to high noise on the raw force sensor data, we process the signal with a 5$^{th}$ order Butterworth filter designed for the \SI{800}{\hertz} measurement frequency with a cutoff of \SI{5}{\hertz}. The filtered signal is plotted, as well as the raw data, which extends beyond the plot range. In general, force magnitudes match the ground truth measurements well. \Ac{RMSE} of the force estimate compared to filtered ground truth is shown in \cref{tab:gt_error}. In the direction of contact, an error of \SI{1.62}{\newton} is due largely to a slow response to a step force, which is influenced by the chosen $\bm{K}_I$. Further tuning of $\bm{K}_I$ could improve fidelity of the estimator without causing instability. Torque error was not evaluated due to the lack of a fixed connection to the sensor, and remains as future work. Experimental results have demonstrated that the system is able to interact for inspection tasks with the chosen parameters.

\begin{table}
\centering
\def\arraystretch{1.2}
\vspace{5mm}
\begin{tabular}{c c c c c c}
\hline
$x$ [m] & $y$ [m] & $z$ [m] & roll [rad] & pitch [rad] & yaw [rad] \\
0.0123 & 0.0106 & 0.0093 & 0.0577 & 0.0488 & 0.0129 \\
\hline
\end{tabular}
\begin{tabular}{c c c}
\vspace{-3mm} \\
\hline
$f_{x}$ [N] & $f_{y}$ [N] & $f_{z}$ [N] \\
1.62 & 0.91 & 0.93 \\
\hline
\end{tabular}
\caption{\Ac{RMSE} of pose and force estimates.}
\label{tab:gt_error}
\end{table}

\section{Conclusion}
\label{sec:conclusion}

A practical system for high force and torque aerial contact applications has been achieved in the form of a novel tilt-rotor omnidirectional aerial manipulation platform. Various experiments demonstrate the system's ability to track full pose trajectories with interaction while rejecting disturbances using an impedance controller with selective apparent inertia. We further validate the system's ability to act as a tool for contact-based inspection through experiments with a \ac{NDT} sensor.

This work forms the basis needed for autonomous contact inspection of complete civil structures. Several open issues of the current system motivate future development. Wall and ground effects are not differentiated from contact forces in the momentum-based wrench estimator and may cause difficulty controlling interaction forces. Though counteracted by the impedance controller, estimation of wind disturbance could improve overall system performance. Drift in position and yaw over time of the \ac{VI} state estimator prompts the additional integration of an absolute localization source, such as a \ac{GPS}, a total station theodolite or feature-based localization. Future extensions of this work will integrate the control approach with mapping, and generate coverage trajectories for complete structural evaluation.

\section*{Acknowledgements}
This work was supported by funding from ETH Research Grants, the National Center of Competence in Research (NCCR) on Digital Fabrication, NCCR Robotics, and Armasuisse Science and Technology.

%% Use plainnat to work nicely with natbib.
\bibliographystyle{plainnat}
\balance
\bibliography{main}

\end{document}